\documentclass{ieeeaccess}
\usepackage{cite}
\usepackage{amsmath,amssymb,amsfonts}
\usepackage{algorithmic}
\usepackage{graphicx}
\usepackage{textcomp}
\def\BibTeX{{\rm B\kern-.05em{\sc i\kern-.025em b}\kern-.08em
    T\kern-.1667em\lower.7ex\hbox{E}\kern-.125emX}}
    
\usepackage{rawfonts, graphicx, amsmath, amssymb, amsfonts, algorithmic, textcomp, caption, subcaption, dirtree, enumitem, xcolor, balance, footnote, multicol}   
\definecolor{mycolor}{gray}{0.97}

\usepackage{ulem}
\usepackage{hyperref}

\begin{document}

\title{Machine-Generated Hierarchical Structure of Human Activities to Reveal How Machines Think}
\author{\uppercase{Mahsun Altin}\authorrefmark{1}, 
\uppercase{Furkan Gürsoy\authorrefmark{2},  and Lina Xu}\authorrefmark{3}
}
\address[1]{Dept. of Computer Engineering, Abdullah Gül University, Kayseri, Turkey (e-mail: mahsun.altin@agu.edu.tr)}
\address[2]{Dept. of Management Information Systems, Boğaziçi University, Istanbul, Turkey (e-mail: furkan.gursoy@boun.edu.tr)}
\address[3]{Dept. of Computer Science, University College Dublin, Dublin, Ireland (e-mail: lina.xu@ucd.ie)}
\tfootnote{DOI: \textcolor{blue}{\url{https://doi.org/10.1109/ACCESS.2021.3053084}}}

\corresp{Corresponding author: Furkan Gürsoy (e-mail: furkan.gursoy@boun.edu.tr).}

\begin{abstract}
Deep-learning based computer vision models have proved themselves to be ground-breaking approaches to human activity recognition (HAR). However, most existing works are dedicated to improve the prediction accuracy through either creating new model architectures, increasing model complexity, or refining model parameters by training on larger datasets.
Here, we propose an alternative idea, differing from existing work, to increase model accuracy and also to shape model predictions to align with human understandings through automatically creating higher-level summarizing labels for similar groups of human activities. 
First, we argue the importance and feasibility of constructing a hierarchical labeling system for human activity recognition. 
Then, we utilize the predictions of a black box HAR model to identify similarities between different activities. Finally, we tailor hierarchical clustering methods to automatically generate hierarchical trees of activities and conduct experiments. 
In this system, the activity labels on the same level will have a designed magnitude of accuracy and reflect a specific amount of activity details. This strategy enables a trade-off between the extent of the details in the recognized activity and the user privacy by masking some sensitive predictions; and also provides possibilities for the use of formerly prohibited invasive models in privacy-concerned scenarios. Since the hierarchy is generated from the machine's perspective, the predictions at the upper levels provide better accuracy, which is especially useful when there are too detailed labels in the training set that are rather trivial to the final prediction goal. Moreover, the analysis of the structure of these trees can reveal the biases in the prediction model and guide future data collection strategies.
\end{abstract}


\titlepgskip=-26pt

\maketitle
\section{Introduction}\label{sec:intro}
\PARstart{V}{ideo} processing attracts a huge demand and interest from both academia and industry due to its ability to unlock the intelligence in many domains such as surveillance, gaming,  autonomous vehicles, medical imaging, human activity recognition, and alike. 
Therefore, video processing and analysis have become a predominant research topic in the artificial intelligence (AI) and machine learning (ML) field.
The performance of video processing tasks has been brought to a new level with the utilization of deep learning (DL).

In existing ML models for video processing and labeling approaches, we have problems such as (i) highly correlated labels and low accuracy for activities involving richer context (e.g., labels with very specific definitions when the number of labels is high) (ii) too detailed prediction results that can violate privacy (iii) trivial labels that have no meaning for a specific use scenario.

First, existing ML models increasingly try to find specific labels for a larger number of activities. 
Their accuracy may become lower since it will be more challenging to distinguish between different but similar activities. 
Without understanding the correlation between the observed data, more data collection can be meaningless\cite{bookwhy}. 
Besides, without understanding the fundamentals of the model (often referred to as the machine), it is hard to redesign the model structure. 
Given these issues, the explainable AI approach \cite{gunning2017explainable, xaiadadi, DBLP:journals/corr/abs-1712-09923} aims to understand how the machine thinks while possibly using such understanding to improve prediction accuracy or usefulness in real-world scenarios with various constraints.
Most existing work has been focusing on training data set and tuning fine models, while limited attention has been given to the post-processing related to the model predictions results.  

Second, privacy has become a highly concerning issue in many human-computer systems. In the context of monitoring, privacy has been a controversial and focal point, especially with the enforcement of certain laws such as the General Data Protection Regulation (GDPR). 
The massive video image collections and the intelligence of processing ability have already brought many scary thoughts to the public.
For example, in the healthcare domain, being able to predict human's daily activities of living (ADL) and perform auto logging will greatly help to build an evidence-based medicine system. 
However, ADL classification through video can be extremely invasive without information filter control.
With existing ML models, it is hard to control and balance between the prediction accuracy and privacy. 
Hence, there should be a way to trade off between the details presented from the video analysis results and privacy while retaining the effectiveness of the system and satisfying the privacy expectations of people.

Third, with the increasing number of possible activity labels, some of those labels have no use in many real-world applications. For instance, the label \textit{alligator wrestling} from \textit{Kinetics-600} dataset will not be relevant in most use cases. Hence, such models and their results should be processed with an appropriate strategy that accounts for the trivial labels. 

\begin{figure*}[h]
    \centering
    \begin{subfigure}[b]{0.48\textwidth}
       \includegraphics[width=\textwidth]{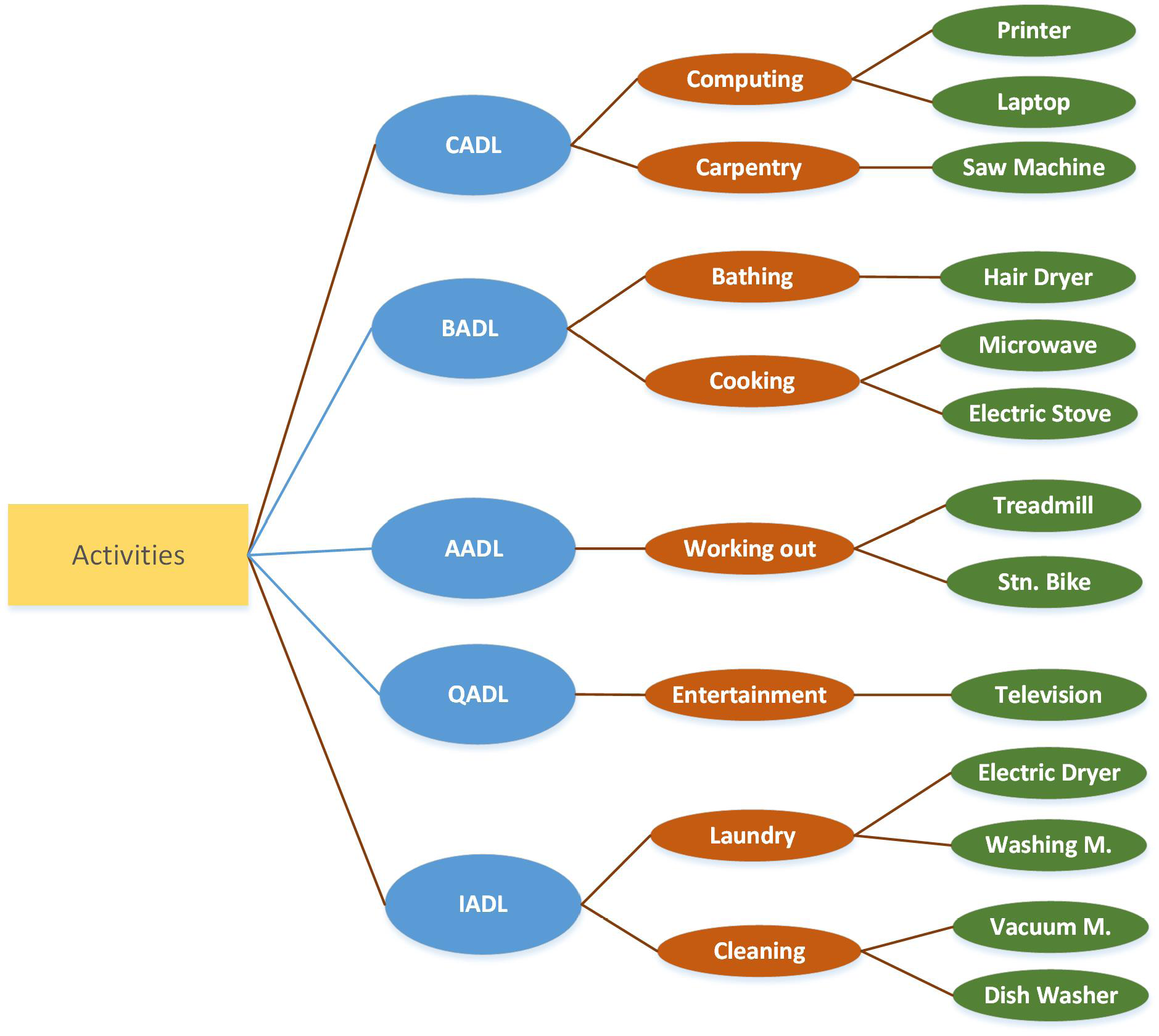}
       \caption{Taxonomy Classification: Basic ADL (BADL), Instrumental ADL (IADL), Extended ADL (EADL), and Advanced ADL (AADL) (Adapted from \cite{tox_yassine_2015}). }
       \label{subfig_tox}
    \end{subfigure}
    \hfill
    \begin{subfigure}[b]{0.48\textwidth}
       \includegraphics[width=\textwidth]{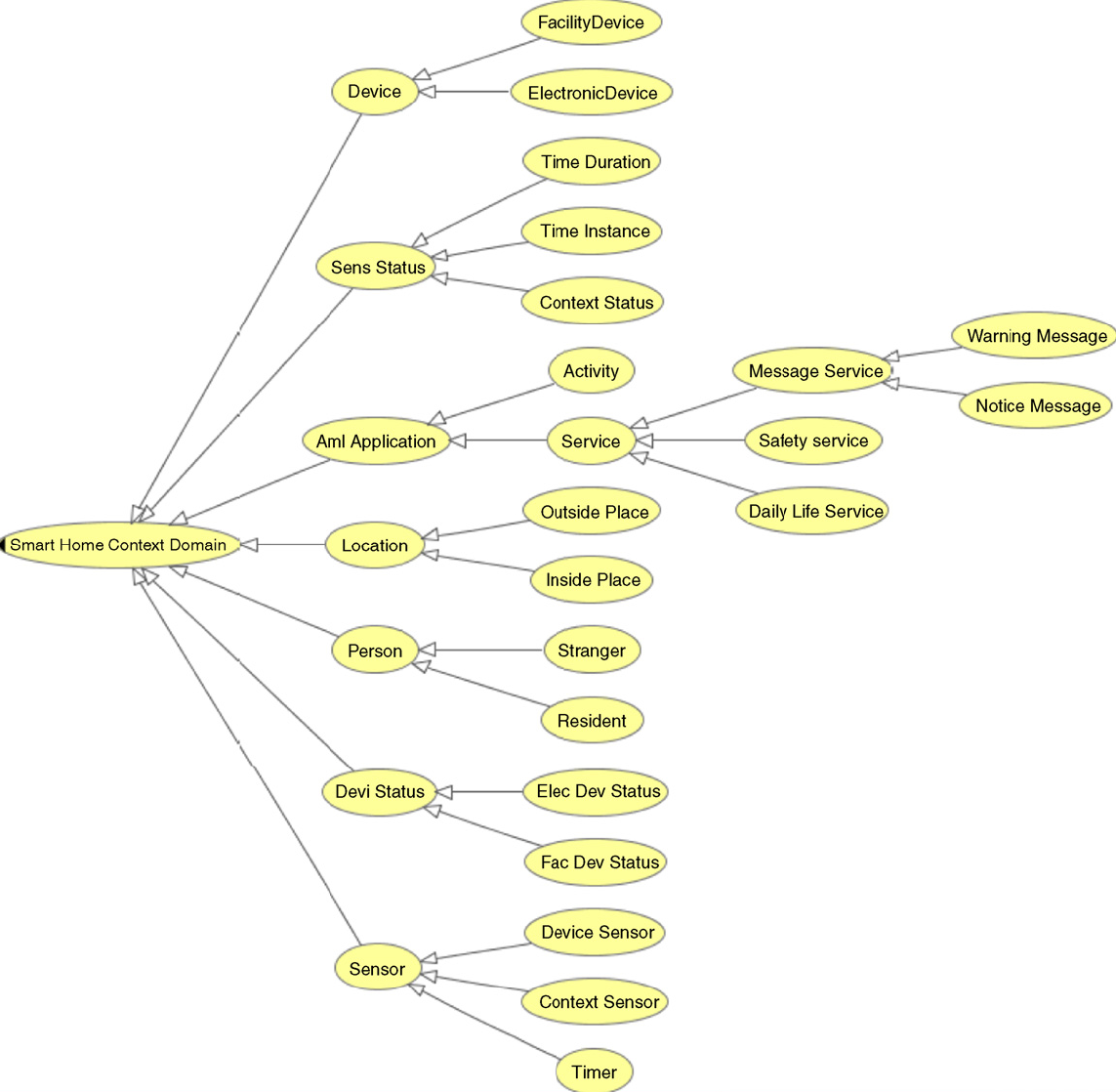}
       \caption{ Web Ontology Language (OWL) based Ontology for ADL Recognition in Smart Homes (Adapted from \cite{onto_2014}). }
       \label{subfig_onto}
    \end{subfigure}
    \caption{Examples of Existing Approaches for ADL Understanding and Classification}
    \label{fig_classfication}
\end{figure*} 

The traditional way of understanding structure and relationships among human activities is based on taxonomy or ontology knowledge graphs, as visually exemplified in Figure~\ref{fig_classfication}. 
However, in existing approaches, the structure of these taxonomies (i.e., activity trees) are generated from the human's perspective.
Yet, if we want to apply ML models for analyzing videos to identify ADL and evaluate the models' intelligence in order to redesign the model, it is also important to understand the classification logic from the machine's perspective.
This is essential to improve and tailor the ML models' performance in a clear direction.

\begin{figure}[h]
    \centering
    \includegraphics[width=0.5\textwidth]{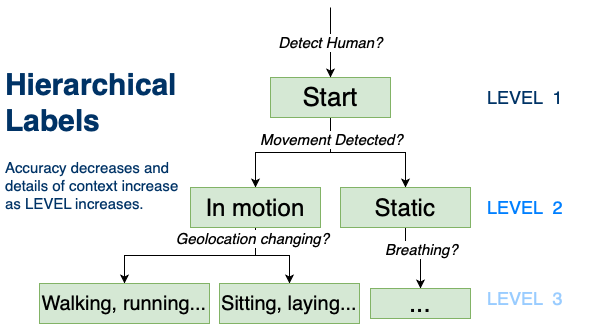}
    \caption{Analysis Supporting a Hierarchical Labeling System}
    \label{fig_label_hire}
\end{figure}

In order to address the above-stated problems, we present a hierarchical labeling system for human activity recognition. 
Such a system is inspired from the machine's perspective and aims to provide adjustable accuracy and privacy for different use scenarios. 
The concept of the system is illustrated in Figure~\ref{fig_label_hire}.
Each level has a designed accuracy range and context richness. 
Based on the use case requirements on privacy, the system can be set at a specific level.
For example, in an edge-cloud configuration, the data collection and analysis can be done on the edge and only summarized results can be transmitted to the cloud so that the private information is not seen or stored by the system operator. 
Since the hierarchical structure is generated through observations from the machine's perspective, the prediction accuracy will increase as LEVEL moves from bottom to up. Moreover, the similarities between labels as understood by the machine will be revealed by the hierarchical tree, which in turn enables model understanding and may guide future data collection strategies.

To the best of knowledge, this is the first study to propose a fully machine-generated human activity hierarchy to overcome above-stated problems and reveal the biases of a HAR machine, and explore how it sees the world; eventually allowing and guiding researchers and practitioners to design appropriate mechanisms in real-world use scenarios and also to improve the model accuracy. In the next section, we provide a brief review of methods and datasets for human activity recognition, present our motivation, and discuss the feasibility of our approach. In Section 3, we describe our methods for automatically generating the hierarchical trees and demonstrate our strategy utilizing a chosen deep learning model and several public datasets. We present experimental results and provide quantitative and qualitative analyses in Section 4. Finally, we present our final remarks and a brief discussion on future directions.

\section{Background}

\subsection{Related Work}

Human activity recognition aims to analyze and detect different human actions from various data sources such as sensor data or multimedia data. It finds application areas in security and surveillance, healthcare, autonomous driving, robotics, smart home, entertainment, and others.
Generally speaking, the literature on HAR is clustered around two topics: the more traditional sensor-based HAR and vision-based HAR. For vision-based HAR, earlier methods include more traditional machine learning algorithms such as decision trees or support vector machines which require hand-engineered features. Recently, deep learning-based models gained popularity since they do not require a time-consuming manual feature generation process and show superior performance.
In particular, sequential learning methods (e.g., long-short term memory) and convolutional neural networks play a key role in the architecture of the state-of-the-art deep learning-based HAR models \cite{seqcnn}.

\begin{figure}[h]
    \centering
    \includegraphics [width=1\linewidth]{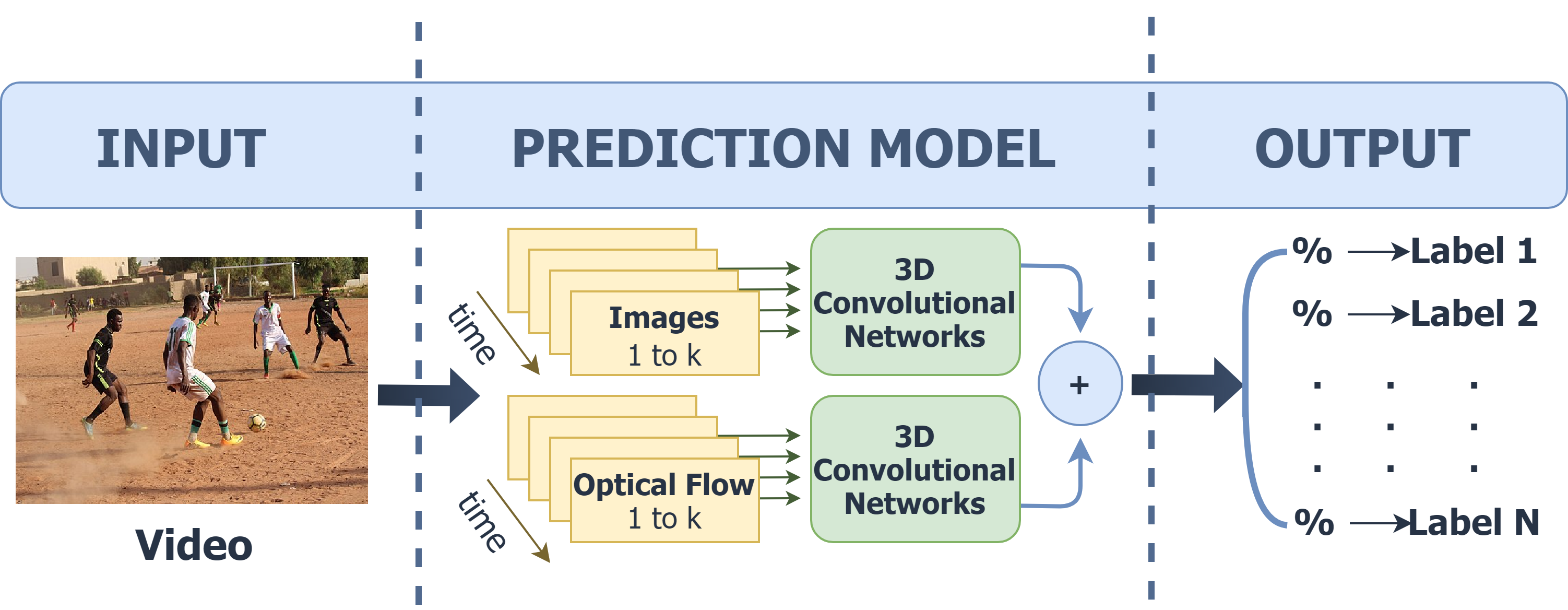}
    \caption{Two-Stream Inflated 3D ConvNet Model, Input and Output.}
    \label{fig:i3d}
\end{figure}

Providing a comprehensive review of the developments in and the current state of HAR research is not possible given the limited space and is beyond the scope of this work. We refer interested readers to the recent comprehensive survey paper by Dang et al. \cite{MINHDANG2020107561} that provides a relatively concise review of hundreds of relevant studies among which exist more than 10 other HAR survey papers. However, here, we briefly describe the model used in this study. Two-Stream Inflated 3D Convnet (\textit{I3D}) \cite{DBLP:journals/corr/CarreiraZ17} is a state-of-the-art deep learning-based computer vision model used in human action recognition. As illustrated in Figure~\ref{fig:i3d}, taking an RGB video as the input, the model produces a series of probabilities corresponding to the labels. The label with the highest probability can be considered as the prediction result for the video. Alternatively, the top $k$ results can be considered together as top predictions.

Many human activity datasets of different formats and coverage have been employed in the literature. It is beyond the scope of this work to list and present datasets. There are studies that specifically review and compare HAR datasets in the literature \cite{datasurvey1, datasurvey2, datasurvey3}. Instead, here, we only present a summary of several public, popular, and relevant video datasets in Table~\ref{tab:dataset}.
\textit{Kinetics} has been first published as \textit{Kinetics-400} \cite{DBLP:journals/corr/CarreiraZ17}. It has 400 different classes and $306,245$ videos all with a length of 10 seconds.
\textit{Kinetics-600} extended it to cover 600 classes $495,547$ videos \cite{DBLP:journals/corr/abs-1808-01340}. Recently, \textit{Kinetics-700} was made available \cite{DBLP:journals/corr/abs-1907-06987}.
All of the video labels in Kinetics dataset are homogeneous and no hierarchical labels are considered.
Kuchne et al. \cite{Kuehne11} published \textit{HMDB51} which covers 51 different human activities with $6,766$ videos, with a minimum $1$ second long length. 
All the videos are also labeled with meta labels, which can be used to create higher classes among those 51 classes. Soomro et al. 
\cite{ucf101} published \textit{UCF101} which covers 101 different classes and $13,320$ videos.
It applied a simplified approach of ontology where the action categories can be divided into five types: human-object interaction, body-motion only, human-human interaction, playing musical instruments, and sports. 
\textit{ActivityNet} was originally published as \textit{ActivityNet-100} \cite{Heilbron_2015_CVPR} and later updated as \textit{ActivityNet-200}, containing $849$ hours long videos in total.
It has organized the labels into a relatively richer hierarchical tree through the expert taxonomy approach. Similarly, \textit{FCVID} \cite{FCVID} categorizes activities in 11 high-level groups.
Lastly, \textit{COIN} dataset \cite{tang_ding_rao_zheng_zhang_zhao_lu_zhou_2019} employs a similar taxonomy classification with 3-stages (domain, task, and step) and contains $11,827$ instructional videos related to 180 different tasks in 12 domains.

\begin{table*}[h]
    \centering
    \begin{tabular}{|c|c|c|c|c|}
          \hline
          \textbf{Name} & \textbf{\# of Videos} & \textbf{\# of Classes} & \textbf{Maintained By} & \textbf{Hierarchical Label} \\
          \hline
          Kinetics-400. & 306,245  & 400  & Deepmind  & No \\ 
          \hline
          Kinetics-600. & 495,547  & 600  & Deepmind  & No   \\ 
          \hline
          HMDB51 & 6,766  & 51  & Serre-Lab  &  Yes, meta labels  \\ 
          \hline
          UCF101 & 13,320  & 101  & U. of Central Florida  & Yes, simplified ontology  \\ 
          \hline
          ActivityNet 200  & 27,801  & 203  & ActivityNet  & Yes, taxonomy  \\ 
          \hline
          FCVID & 91,223 & 239 & Fudan University & Yes, taxonomy  \\ 
          \hline
          COIN & 11,827 & 180 & Tsinghua University & Yes, taxonomy  \\ 
          \hline
    \end{tabular}
    \caption{Open Video Datasets}
    \label{tab:dataset}
    \vspace{3mm}
\end{table*}
As previously illustrated in Section~\ref{sec:intro}, existing studies focusing on human activity hierarchies employs expert generated hierarchies such as in \cite{tox_yassine_2015, onto_2014, adl_ontology}. Long et al. \cite{long_mettes_shen_snoek_2020} utilizes action hierarchies represented in a hyperbolic space rather than simpler tree representations, where hierarchies are based on some external expert guidelines from \textit{ActivityNet}. Gaidon et al. \cite{gaidon_harchaoui_schmid_2013} represent individual videos as a hierarchy of activities based on spatio-temporal trajectories but do not provide a global taxonomy of activities. Our strategy differs from these and brings a novel approach to understand the behavior and bias of an HAR machine and leverage the power of human intelligence to improve it.


\subsection{Motivation and Feasibility}

The ultimate goal of AI is to train the machine to think in a consistent way with how a human thinks, particularly in tasks where humans are good at such as human activity recognition.
Unfortunately, we are currently far from that and there is still a large gap between how the machines think and how we humans think. 
In order to reduce this gap, it is necessary to assess where the machine's intelligence is through understanding the analysis results from the machine's perspective. 
This awareness will be a key to enhance the learning model and it can also provide directions for future data collection. 
In order to train the machine and collect more data in an efficient and effective manner, one needs to understand the correlation between the existing videos and address the limitations and boundaries. 
Most existing models are mainly concerned with the accuracy for each label and video.
We think that introducing a hierarchical prediction system can better serve the goal.
Moreover, one huge concern currently about video processing is privacy invasion.  
Limited work concerned itself with the privacy problem when designing an analyzing/predicting machine learning model for ADL recognition. 

In order to modify existing deep learning models to output prediction results through a hierarchical labeling structure, firstly we need to analyze the current prediction results and prove that there is such a possibility.
This requires us to understand the human activities from the machine's perspective, rather than from our human intuition.
The machine learns about the world through the dataset that is fed to them, more specifically, videos and the labels that are associated with them. 
The model might be biased and contradicting to our intuition due to the explicit or implicit limitations of the dataset. 
For example, from the perspective of a human or a real-world use case, we may think videos of \textit{brushing teeth} and \textit{flossing} are quite similar. 
However, an ML model may classify them quite differently and identify that \textit{flossing} has a higher similarity to \textit{putting makeup}.
This may simply because the training dataset does not have enough number of diverse videos for these labels, hence, the model cannot identify this activity precisely. 

We have examined several existing video analysis deep learning models and discover that the total number of labels are huge, ranging to more than 600 labels. 
Some of the labels are revealing great details of human activities. 
In many scenarios, too detailed labeling is not necessary and also invasive to privacy. 
For example, due to privacy issues, there would be different requirements for the analysis results and the level of details. 
Besides, too detailed labels make training those models become difficult since sourcing appropriately rich and diverse training data is challenging.
Due to the limited amount of training data, the prediction accuracy for individual labels can be extremely low, which provides little meaning. 
Some of the labels have high correlations, which makes prediction even more difficult and the ML model can be easily confused. 
Some of the labels are mostly trivial and make little sense in reality, especially in daily activities monitoring. 
To summarize, there are several existing problems accosted with existing video processing  deep learning models:
\begin{itemize}
    \item Too many labels with only homogeneous organizing   
    \item Low accuracy for top 1 prediction 
    \item High correlation between many labels
    \item Too detailed labeling invading privacy
\end{itemize}

The Two-Stream I3D model trained on \textit{ImageNet}\cite{deng2009imagenet} and \textit{Kinetics} dataset yields the following performance on \textit{Kinetics-400}.  The accuracy for the Top 1 prediction is 74.2\% whereas the accuracy for the sum of top 5 predictions can reach 91.3\%.
Some activities with more details/context involved are harder to predict than the ones with less detail. 
For example, \textit{tango dancing} is harder to identify comparing with \textit{jogging}.
Furthermore, \textit{tango dancing} is reflecting more details related to a scenario than \textit{dancing}. 

This motivates us to investigate into the similarity of the labels and build a hierarchical tree structure of them by ML techniques. 
The depth of the tree indicates the accuracy and level of details of the prediction result to fulfill a system's privacy and accuracy requirements.
In order to construct such a hierarchical tree for ADL, we need to understand the correlations between the labels and this understanding needs to be from the machine's perspective. 
Through this way, we can discover which parts of the machine's thinking are contradicting to our human common sense. 
Therefore we can fill the gap via specifically directed future improvements. 
Then, for privacy concerns, we need to understand the connection between privacy and accuracy. 
We can choose a method to cluster activity labels into a hierarchical structure to enable the high prediction accuracy at desired, privacy-preserving levels.


\section{Hierarchical Labelling Solution}

The methodology in this study is designed to generate a hierarchy labeling tree for human activities and can be conducted in two consecutive steps. 
First, similarity measures are devised to quantify the resemblance between different activities through their co-occurrences in the predictions of a video classification model.
This resemblance is measured and assessed from the machine's perspective rather than the human's. 
Then, considering the quantified similarities between activities as the distance metric, a set of various hierarchical clustering techniques are applied to automatically generate the activity label trees.
Through analyzing the generated tree from the human's perspective, we can gain insights into the model and the dataset. 

\subsection{Similarity Space}
In order to construct the similarity dataset, we have followed the following procedures.

\textbf{Composed Dataset:} A collective dataset with $15,938$ videos are formed from public datasets including \textit{UCF101}, \textit{HMDB51}, and \textit{Kinetics-600}. Herein we refer this dataset as the \textit{ComposedDS}.
The whole \textit{UCF101} dataset which consists of $12,435$ videos is taken. 
$3456$ videos are chosen randomly from \textit{HMDB51}. However, the selected videos from the two datasets did not have any videos that resembled the 6 of the 600 activities from \textit{Kinetics}, which hints at the discrepancy between different datasets. 
To ensure that all labels show up at least once in the top-5 predictions of the \textit{I3D} model, $47$ videos of these 6 activities are taken from the test set of \textit{Kinetics-600}.

\textbf{Video classifier:} The \textit{I3D} model pre-trained on the training set of \textit{Kinetics-600} is used to make predictions for each of video from \textit{ComposedDS}. 
The model has an output vector with 600 values through a softmax function.
Each value in the vector represents the predicted probability of the corresponding label. 
The prediction results for all the videos from \textit{ComposedDS} are recorded and used to form a generated dataset with a size of [$15938$, $600$], herein named \textit{GeneratedDS600}.
It is important to note that the \textit{I3D} model was not trained on any of the videos from \textit{ComposedDS} and the original ground-truth labels of \textit{ComposedDS} differs from that of \textit{Kinetics-600}.
Therefore, the dataset that the \textit{I3D} model is trained on does not overlap with the dataset that is used to generate \textit{GeneratedDS}.
For each video, the top five predictions of \textit{ComposedDS} are recorded as an individual set, named \textit{GeneratedDS5} with a size of [$15938$, $5$]. 
For an example video, the top five predictions are stored in \textit{GeneratedDS5} as shown in Table~\ref{tab:topfive}, originally followed by No. 6 \textit{passing American football (AF), in game} with a probability of 0.02.
\begin{table}[b]
    \centering
    \begin{tabular}{l||c}
          \textbf{Predicting Label} & \textbf{Predicting Score} (Sorted) \\
          \hline
          shooting goal (soccer)              & 0.58 \\ \hline
          passing soccer ball                 & 0.18 \\ \hline
          playing field hockey                & 0.06  \\ \hline
          kicking soccer ball                 & 0.06  \\ \hline
          juggling soccer ball                & 0.04  \\ \hline
    \end{tabular}
    \caption{\textit{I3D} Predictions for a Sample Video}
    \label{tab:topfive}
\end{table}

A really interesting fact we have observed from Table~\ref{tab:topfive} and many other instances from \textit{GeneratedDS5}, is that most of the time, the activities that the machine thinks are correlated, are also correlated from our human's understanding. 
This gives us a good start. 
\textit{I3D} model is recognizing the world similarly as a human to a certain degree. 
However, it is not always the case. 
As in Table~\ref{tab:topfive}, the machine thinks that \textit{playing field hockey} is similar to other soccer activities and ranks better than few other soccer activities in this case.
This mistake is not dramatically disappointing since the probabilities associated with those labels are not high.
With a simple analogy, if a child has made this mistake, blindly forcing them to remember this activity is not an efficient way of education.
Finding the correlation between the activities and teaching the child to learn the similarities and the distinctions is a better approach.    
Therefore, teaching the child to distinguish a soccer field from the hockey field may avoid this mistake in the future. The same is true when training machines.

\textbf{Similarity calculation:} 
To teach the machine to be more intelligent, we want to adopt a more systematic way rather than just randomly collecting more data and perform more training.
We need to understand why the machine made this mistake.
Like teaching a child, firstly, we need to find out, which labels that the machine thinks are highly similar. 
We propose a similarity measure based on the element co-occurrences in \textit{GeneratedDS5}. 
We propose to use two different measures to quantify the extent of co-occurrence: \textit{confidence} \cite{agrawal} (Eq. \ref{eq:consim}) and \textit{lift} \cite{bayardo} (Eq. \ref{eq:lifsim}). 

\begin{equation}
\label{eq:consim}
C_{ij} = P(i|j) = \frac{P(\{i,j\})}{P(j)}
\end{equation}

\begin{equation}
\label{eq:lifsim}
L_{ij} = \frac{P(i| j)}{P(i)} = \frac{P(\{i,j\})}{P(i)P(j)}
\end{equation}

\textit{Confidence} $C_{ij}$ measures the probability of $i$ being in the set given that $j$ already exists. $C_{ij}$ can range from 0 to 1 and a higher value indicates a greater similarity between the activity $i$ and $j$. It is an asymmetric measure; thus, $C_{ij} \neq C_{ji}$ usually. 
Moreover, if $i$ is a popular activity in the dataset, it tends to co-occur more with others, resulting in having higher \textit{confidence} values.

\textit{Lift} $L_{ij}$ corrects $C_{ij}$ for the overall popularity of $i$ over all sets. When it equals $1$, it indicates the complete independence of events $i$ and $j$, i.e., the extent of their co-occurrences is by random chance. Like \textit{confidence}, the higher values indicates greater similarity. Unlike the former which is limited to the $[0,1]$ range, it can take any non-negative value and it is a symmetric metric; thus, $L_{ij}  = L_{ji}$.

We can consider that high \textit{confidence} or \textit{lift} value indicate high similarity between two activities from the machine's perspective. 
However, we are also interested in the similarity structure among all activities. 
We use \textit{confidence} and \textit{lift} as the distance metrics to cluster activities into subgroups.
The information for the subgroups is important for testing a machine and further training. 
Using the same analogy, when asked to describe an activity, a child might say \textit{playing soccer}. 
This may be a wrong answer, but you can still get the general concept that people are playing a certain sport. 
If people are not \text{playing soccer} but are \text{playing sports}, then you can still trust the explanation of the child at a certain level.
If people are not even \text{playing sports}, then you cannot rely on the responses of the child at all. 
This hints that there are more fundamental issues about how the child is taught to distinguish different activities.
Likewise, in the machine intelligence case, we want to know what is the probability associated with \text{playing sports} and what is the probability associated with \text{playing soccer}. 
Therefore, hierarchical clustering for activities is a good strategy to assess the intelligence of a machine and also can help with teaching a machine to learn.

For hierarchical clustering, the distance between $i$ and $j$ must be symmetric, i.e., $D_{ij} = D_{ji}$. \textit{Lift} is symmetric whereas \textit{confidence} is not. 
We define the distance $D_{ij}$ between two activities $i$ and $j$ in two alternative ways based on \textit{confidence} and \textit{lift} as shown in Equation \ref{eq:simcon} and \ref{eq:simlift} respectively and ensure that both are generalised to the range $[0, 1]$ for consistency.

\begin{equation}
\label{eq:simcon}
D_{ij}^C = = 1 - \sqrt{C_{ij} C_{ji}}
\end{equation}

\begin{equation}
\label{eq:simlift}
D_{ij}^L = 1 - \frac{L_{ij} - \min(L)}{\max(L) - \min(L)}
\end{equation}

In obtaining a single symmetric measure from two asymmetric \textit{confidence} measures, geometric mean is chosen over arithmetic mean on purpose to award reciprocal similarities where activity $i$ co-occurs with activity $j$, and $j$ also co-occurs with $i$. 
The measure obtained by geometric mean is referred to as the \textit{cosine} measure whereas the measure that can be obtained by the arithmetic mean is referred to as \textit{Kulczynski} measure. 
\textit{Cosine} measure is \textit{null-invariant}, i.e., it is not affected by the sets that do not contain any of the examined elements (i.e., null sets). 
On the other hand, \textit{lift} is sensitive to the number of such null sets, which
\cite[Chapter~6]{han2011data} provides a comparative discussion on different measures in the context of frequent itemset mining. 
As a demonstrating example, Table~\ref{tab:basketball} lists each activity that is a member in at least 5\% of all sets where \textit{playing basketball} is also a member, along with the values calculated by the similarity measures employed in this study.

\begin{table}[h]
\centering
\begin{tabular}{l||l|l|l|l}
\hline
\hspace{13 mm}  $i$  & $P (i | j)$ & $P(j | i )$ & $C_{ij}$ & $L_{ij}$ \\ \hline
shooting basketball                               & 0.76        & 0.87        & 0.81     & 23.5    \\ \hline
dunking basketball                                & 0.64        & 0.87        & 0.74     & 23.5    \\ \hline
dribbling basketball                              & 0.54        & 0.82        & 0.67     & 22.1    \\ \hline
playing netball                                   & 0.33        & 0.58        & 0.44     & 15.5    \\ \hline
playing volleyball                                & 0.20        & 0.63        & 0.36     & 17.0    \\ \hline
juggling soccer ball                              & 0.19        & 0.30        & 0.24     & 8.0     \\ \hline
passing AF, not in game                            & 0.14        & 0.51        & 0.27     & 13.8    \\ \hline
throwing ball                & 0.11        & 0.25        & 0.17     & 6.7     \\ 
(not baseball or AF)  & & & &\\ \hline
dodgeball                                         & 0.08        & 0.34        & 0.16     & 9.3     \\ \hline
shooting goal (soccer)                            & 0.07        & 0.07        & 0.07     & 2.0     \\ \hline
playing tennis                                    & 0.06        & 0.05        & 0.05     & 1.4     \\ \hline
high kick                                         & 0.05        & 0.06        & 0.05     & 1.5     \\ \hline
\end{tabular}
\caption{Top Items Co-occurring with $j  : $ \textit{playing basketball}}
\label{tab:basketball}
\end{table}

\subsection{Hierarchical Clustering}

Clustering is the process of creating multiple groups (i.e., clusters) from the given objects such that the objects are similar to those in the same clusters and dissimilar to those in different clusters. 
We employ hierarchical clustering (HC) methods to obtain a tree-like structure where the number of clusters can be determined post-hoc and the interpretation is richer.

Agglomerative hierarchical clustering works by merging the most similar objects in a tree-like structure from leaf nodes to the root node in a bottom-up manner. 
In the beginning, it creates the same number of clusters as the number of objects where each object is a cluster of its own, called as a singleton. 
Then, at each step, it identifies the most similar two clusters and merges them into one cluster. 
This process iterates until all clusters are merged into the single highest-level cluster. 
The order in which clusters are merged with each other, hence, results in the hierarchical tree structure.

There exist multiple methods for calculating the distance between two clusters. Given two clusters $u$ and $v$, \textit{single-linkage} uses the shortest distance between any pair of objects from $u$ and $v$, \textit{complete linkage} uses the longest distance between any pair of objects from $u$ and $v$, \textit{average-linkage} \cite{sokal1958statistical} uses the average distance between all pairs of objects from $u$ and $v$. Given that $s$ and $t$ is merged to form cluster $u$,  \textit{weighted-linkage} \cite{sokal1958statistical} uses the average of distances between $s$ and $v$, and $t$ and $v$ to find the distance between $u$ and $v$. On the other hand, \textit{Ward's} method \cite{ward1963hierarchical} merges two clusters which minimizes the total within-cluster variance at each step. According to \cite{hands1987monte} and \cite{timm2002applied}, \textit{Ward's} and \textit{complete-linkage} methods tends to generate a more balanced tree with approximately equal cluster sizes.

600 activity labels $i, j, \dots$ and the pairwise distances between them $D_{ij}, \dots$ are fed as input to the above-mentioned set of hierarchical clustering methods. In line with its goals and underlying mechanisms, the HC process is expected to reveal the hidden hierarchy of human activities. Moreover, HC algorithms utilizing different methods to calculate the distance between clusters would result in different trees which can be compared quantitatively (e.g., the extent it improves prediction accuracy at activity group levels) and qualitatively (e.g., the extent that the shared semantics within the same activity cluster is plausible).

The trees obtained by the HC method can be cut at different depths to divide the data into the desired number of clusters. In this case, the maximum number of clusters is 600 where each label is a cluster of its own and there is practically no clustering. The minimum number of clusters is 1 where all labels are grouped together and there is again practically no clustering. The tree can be cut to obtain $k$ clusters where $1 \leq k \leq 600$.  2670 videos are randomly selected from \textit{Kinetics-600} test set (excluding the 47 videos utilized earlier) and prediction of \textit{I3D} model is compared at different levels. The accuracy at $k=600$ gives the accuracy with respect to the individual activity labels, the accuracy at $1 < k < 600$ gives accuracy with respect to the activity groups at different levels, and $k=1$ gives the accuracy of $100\%$. For instance, when $k=10$, the activities are organized under 10 groups and accuracy is measured based on whether the label predicted by \textit{I3D} is in the same cluster as the ground-truth label is in.


\section{Results}

\subsection{Overview on the accuracy}

\begin{figure*}[h]
\centering
    \includegraphics [width=1\linewidth]{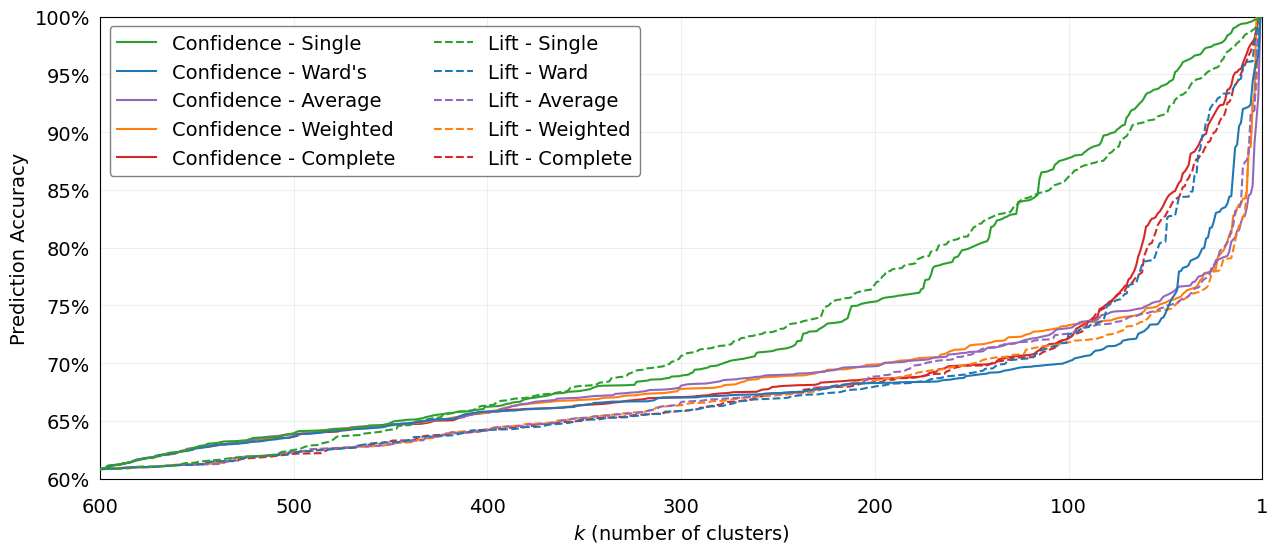}
\caption{Classification Accuracy of \textit{I3D} Model on ADL Hierarchies }
\label{fig:accuracy}  
\end{figure*}

Utilizing the two distance measures and five hierarchical clustering methods, a total of 10 different hierarchies are obtained. Figure~\ref{fig:accuracy} visualises the classification accuracy of \textit{I3D} model on randomly selected 2670 videos from \textit{Kinetics-600} test set. Each line represents an activity hierarchy obtained by a different method, $x$-axis shows the hierarchy levels from bottom (more detailed activities) to top (broader categories), and $y$-axis shows the prediction accuracy at corresponding levels of hierarchy. The predictive accuracy is higher at the upper levels of the hierarchy. This is fitting with our human intuition. It makes the machine's predictions at high levels more trustworthy. For instance, from the original \textit{I3D} model, we only get one lengthy final output vector. Even though the original top 1 prediction from the model is wrong, the upper-level predictions have a better chance to be right. 

In very detailed levels where the average cluster size is up to around 2 (i.e., when $k > 300$ approximately), most methods perform comparably. In this regime characterized by very small-sized clusters of activities, \textit{confidence}-based distance measure slightly outperforms \textit{lift}-based distance measure as shown by the consistently underperforming dashed lines in this region \textendash except for \textit{lift}-based \textit{single-linkage} method (dashed green line) which begins to outperform all other methods at around $k = 400$. For $k < 300$, which is the more interesting and real-world case with more clustering, the two \textit{single-linkage} methods are the only contenders with the \textit{confidence}-based version  outperforming the other once the average cluster size is above 6, that is when $k > 100$. Overall, \textit{single-linkage} method utilizing \textit{confidence}-based distance (hereafter called \textit{Confidence-Single}) performs better than the others when clusters include several items (i.e., $k<100$) or very few items  (i.e., $k>400$)  and performs very close to the method with highest accuracy when $100<k<400$.

When Figure~\ref{fig:accuracy} is evaluated as a whole, we observe that a hierarchical labeling system can dramatically improve the prediction accuracy at upper levels, particularly when the activity hierarchy is discovered with an appropriate and effective clustering method. This property carries importance for prediction systems with a large number and variety of activities, particularly when the information at different levels is necessary for specific use cases. 
Generally, for this to be useful, the clusters should be organized in a meaningful way. 
It should be feasible to label or annotate clusters such that the activity labels are consistent with their respective cluster labels at upper levels.

Next, we closely examine, annotate, and discuss the activity hierarchies generated by \textit{Confidence-Single} which has the highest overall accuracy, and by \textit{Confidence-Ward's} which has a more balanced structure.

\subsection{The Hierarchy of Activities}

\begin{figure}[h]
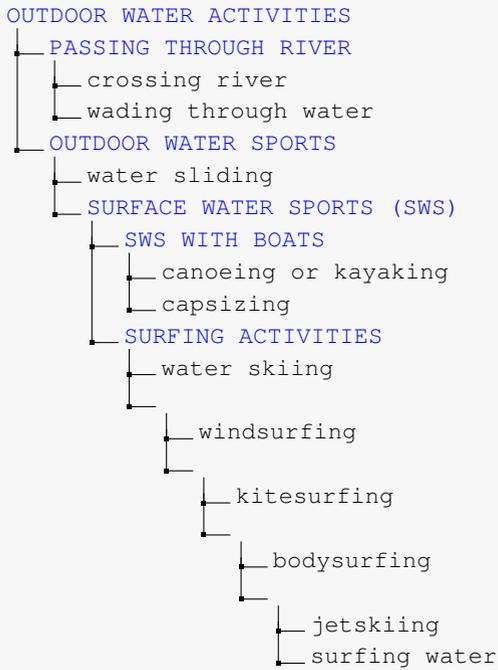

\centering
\colorbox{mycolor}{\parbox{0.485\textwidth}
{\fontsize{9pt}{9pt}\selectfont
\dirtree{%
.1 \textcolor{blue}{\uppercase{outdoor water activities}}.
.2 \textcolor{blue}{\uppercase{passing through river}}. 
.3 crossing river.
.3 wading through water.
.2 \textcolor{blue}{\uppercase{outdoor water sports}}.
.3 water sliding.
.3 \textcolor{blue}{\uppercase{surface water sports (sws)}}. 
.4 \textcolor{blue}{\uppercase{sws with boats}}. 
.5 canoeing or kayaking.
.5 capsizing.
.4 \textcolor{blue}{\uppercase{surfing activities}}. 
.5 water skiing.
.5 . 
.6 windsurfing.
.6 . 
.7 kitesurfing.
.7 . 
.8 bodysurfing.
.8 . 
.9 jetskiing .
.9 surfing water.
}
}}
\caption{Hierarchy of Outdoor Water Activities. }
\label{fig:water}
\end{figure}

\begin{figure}[h]
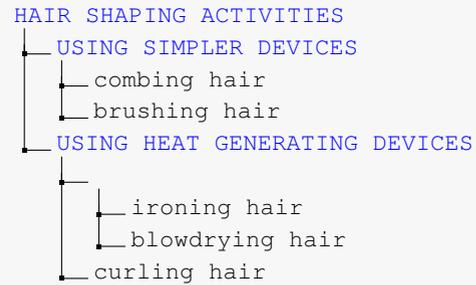

\centering
\colorbox{mycolor}{\parbox{0.485\textwidth}
{\fontsize{9pt}{9pt}\selectfont
\dirtree{%
.1 \textcolor{blue}{\uppercase{hair shaping activities}}.
.2 \textcolor{blue}{\uppercase{using simpler devices}}. 
.3 combing hair.
.3 brushing hair.
.2 \textcolor{blue}{\uppercase{using heat generating devices}}.
.3 . 
    .4 ironing hair.
    .4 blowdrying hair.
.3 curling hair.
}

}}
\caption{Hierarchy of Hair Shaping Activities}
\label{fig:hair}
\end{figure}

\begin{figure}[h]
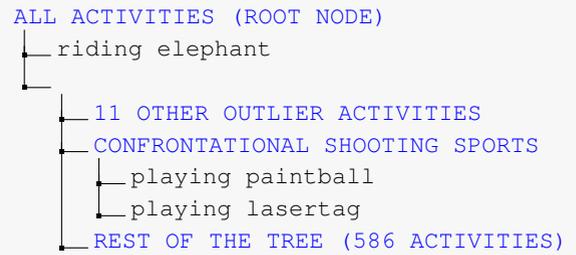

\centering
\colorbox{mycolor}{\parbox{0.485\textwidth}
{\fontsize{9pt}{9pt}\selectfont
\dirtree{%
.1 \textcolor{blue}{\uppercase{all activities (root node)}}.
.2 riding elephant.
.2 .
.3 \textcolor{blue}{\uppercase{11 other outlier activities}}.
.3 \textcolor{blue}{\uppercase{confrontational shooting sports}}.
.4 playing paintball.
.4 playing lasertag.
.3 \textcolor{blue}{\uppercase{rest of the tree (586 activities)}}.
}
}}
\caption{Individual and Small Groups of Outlier Activities  } 
\label{fig:outlier}
\end{figure}

The activity hierarchies generated by the machine and annotated by humans cannot be produced here in full due to space limitations but will be made available online\footnote{https://aiiot.ucd.ie/}. 
Figures~\ref{fig:water}, \ref{fig:hair}, and \ref{fig:outlier} provides some extracts from the tree where original labels are coloured in black and human annotations are written in blue and uppercase letters, and some annotations are omitted for very small clusters.

The hierarchy generated by \textit{Confidence-Single} can be characterized as follows. The major part of the tree consists of clusters whose members are clearly related to each other. For instance, Figure~\ref{fig:water} shows a subtree that brings together various outdoor water sports and Figure~\ref{fig:hair} shows another subtree for hair shaping activities. In general, these figures exemplify the major part where the machine is able to generate intuitive and meaningful hierarchical relations among the activity labels.

The other and minor part, however, consists of the outlier and/or under-identified activities that are not found to be sufficiently similar to most other activities. These activities, individually or in very small groups, are merged into the rest of the tree, i.e., the large tree that consists of the very intuitive groupings as described in the preceding paragraph. Figure~\ref{fig:outlier} exemplifies the characteristics of this part. The singleton \textit{riding elephant} is merged with the rest of the tree, i.e., the cluster of other 599 activities. \textit{Playing paintball} and \textit{playing lasertag} are grouped together to form a meaningful cluster of two before merging with the rest of the tree, i.e., a large cluster of $586$ activities. Usually, activities in this part are not part of meaningful hierarchies that span more than two levels.


Given the limitations of hierarchy generated by \textit{Confidence-Single}, we also investigate the hierarchy generated by \textit{Confidence-Ward's} since it tends to construct clusters of comparable sizes. Unlike the former, there are no activities that are merged to the rest of the tree as individuals or in very small groups. When visually inspected, it appears as a balanced tree with some activities (those activities \textit{Confidence-Single} is not able to place within meaningful hierarchies) are now grouped with relatively similar activities alongside some occasional misplacements. Figure~\ref{fig:wardtree} presents the human-annotated groups of activities based on the hierarchy automatically generated by \textit{Confidence-Ward's}. The binary nature of the tree is omitted for neat visualization. Original activities (in \textit{italic}) are given as examples for each cluster.

\begin{figure*}[h]
\colorbox{mycolor}{\parbox{1.0\textwidth}
{\fontsize{9pt}{9pt}\selectfont
\begin{multicols}{2}

\begin{itemize}[leftmargin=4mm]

    \item Activities involving head
    \begin{itemize}
    
        \item Hair care
        \begin{itemize}
            \item Barber: \textit{shaving head, getting a haircut, washing hair}
            \item Hair shaping: \textit{combing hair, blowdrying hair, curling hair }
        \end{itemize}
        
        \item Facial gestures
        \begin{itemize}
            \item Eye: \textit{winking, raising eyebrows, crossing eyes}
            \item Mouth: \textit{yawning, laughing, sneezing}
        \end{itemize}
        
        \item Eating: \textit{eating spaghetti, eating cake, eating watermelon}
    
    \end{itemize}

    \vspace{2mm}
    \item Food preparation: \textit{making cheese, cooking scallops, rolling pastry, flipping pancake, chopping meat, sharpening knives }

    \vspace{2mm}
    \item Playing musical instruments
    \begin{itemize}
    
        \item Keyboard instruments: \textit{playing organ, playing piano}
        
        \item Wind instruments: \textit{playing clarinet, playing saxophone}
        
        \item Bowed instruments: \textit{playing violing, playing cello}
        
        \item Lutes: \textit{playing ukulele, playing lute, playing guitar}
        
    \end{itemize}

    \vspace{2mm}
    \item Sports I
    \begin{itemize}

        \item Indoor sports
        \begin{itemize}
        
            \item Gym Sports
            \begin{itemize}
                \item Weightless: \textit{yoga, contorting, push up}
                \item Weights: \textit{deadlifting, clean and jerk, bench pressing}
            \end{itemize}
            
            \item Swimming: \textit{swimming backstroke, ice swimming}
        
        \end{itemize}

        \item Outdoor Sports:
        \begin{itemize}
        
            \item Other: \textit{tightrope walking, cartwheeling, somersaulting}
            
            \item Climbing: \textit{climbing a rope, rock climbing, abseiling}
            
            \item In the air: \textit{paragliding, skydiving, parasailing}
            
            \item Skiing: \textit{ski jumping, skiing slalom, skiing mono}
            
            \item Boarding: \textit{longboarding, skateboarding}
           
            \item Cycling: \textit{riding unicycle, falling off bike, jumping bicycle}
            
        \end{itemize}
        
   \end{itemize}
  
    \vspace{2mm}
    \item Sports II
    \begin{itemize}
    
        \item Outdoor water sports
        \begin{itemize}
            \item Involving a boat: \textit{canoeing or kayaking, capsizing}
            \item Surfing: \textit{kitesurfing, windsurfing, water skiing }
        \end{itemize}
    
        \item Involving a relatively large ball
        \begin{itemize}
        \item Football-like: \textit{kicking soccer ball, hurling, tackling}
        \item Indoor: \textit{playing basketball, playing volleyball}
        \end{itemize}
        
        \item Involving a relatively small ball
        \begin{itemize}
            \item Golf: \textit{golf driving, golf putting, golf chipping}
            \item Bat-and-ball games: \textit{playing cricket, hitting baseball}
            \item Throwing: \textit{hammer throw, shot put, throwing discus}
        \end{itemize}
      
        \item Involving no ball
        \begin{itemize}
            \item Jumping: \textit{triple jump, long jump, high jump}
            \item Throwing: \textit{javelin throw}
        \end{itemize}
        
    \end{itemize}

    \item Other
    \begin{itemize}
    
        \item Smoking and Drinking:
        \begin{itemize}
            \item Smoking: \textit{smoking, smoking pipe, smooking hookah}
            \item Drinking: \textit{tasting wine, drinking shots, bartending}
        \end{itemize}

        \item Handcraft: \textit{knitting, making jewelry, bookbinding}
        
        \item Trees: \textit{trimming shrubs, picking fruit, trimming trees}
        
        \item Table games: \textit{playing monopoly, playing chess, playing poker}
        
        \item Dancing I: \textit{breakdancing, dancing macarena, tap dancing}
        
        \item Dancing II: \textit{tango dancing, swing dancing, salsa dancing}
        
        \item Fingers: \textit{clapping, air drumming, twiddling fingers}
        
        \item Ice sports: \textit{playing ice hockey, hockey stop ice skating}
        
        \item Hitting: \textit{punching bag, side kick, punching person (boxing) }
        
        \item Juggling: \textit{spinning poi, juggling balls, juggling fire}

        \item \dots
        
    \end{itemize}
        
\end{itemize}
\end{multicols}
}

} 
\caption{Human Activity Hierarchy}
\label{fig:wardtree}

\end{figure*}

\subsection{Insights on the Datasets and Prediction Model}

The generated hierarchies provide implicit insights into the prediction model and the datasets. Some activity labels are not part of any meaningful multiple-level hierarchies and remain in very small groups or as singletons until near the end when they merge with the rest of the tree. Other activity labels are placed in hierarchies where they do not intuitively belong. This manifests under-identification for those activities. We point out three potential reasons for this problem and suggest appropriate improvements to overcome such under-identification.

First, the example videos that exhibit those activities might be scarce or monotypic in the collected datasets. This results in too few samples to learn the characteristics of these activities and their resemblance to other activities. For instance, \textit{riding elephant} is the top prediction and in the top-5 predictions only for 4 and 19 videos, respectively. To explore whether the scarcity in the dataset is related to under-identification, consider the top 50 activities that are among the first to merge into meaningful and intuitive hierarchies and the bottom 50 activities that are merged individually or in very small groups to the rest of the tree near termination. For the top 50 activities, the median number of videos where the top five predictions include them is 117 whereas the same value for the bottom 50 videos is only 31. The overall median is 78. This indicates that the lack of videos for some activities might be preventing those activities to be placed well in the automatically generated hierarchy.

Second, the set of activities the model is trained for might not be comprehensive enough to cover other activities that are naturally very similar. For instance, consider the activity \textit{milking cow}. The most similar activity might be \textit{milking sheep} or some other farm activity involving \textit{cow}. However, there is almost no other activity in the 600 activities that involve animal husbandry. As a result, \textit{milking cow} could not be placed as part of an intuitive, interesting, or otherwise informative sub-hierarchy by \textit{Confidence-Single}. \textit{Confidence-Ward's} provides only a relatively meaningful hierarchy for \textit{milking cow} where it is first merged with \textit{grooming horse}, and then the two are merged with \textit{bullfighting}.

Third, some activities might have very unique nature that it is hard to group them with other activities. For instance, \textit{playing paintball} and \textit{playing lasertag} form the very intuitive cluster \textit{confrontational shooting sports} (in Figure~\ref{fig:outlier}) but they are not part of a hierarchy with meaningful depth and width since there are not many other activities neither in the model nor in the real world such that they are sufficiently close.

Tracing the individual impacts of these factors on the generated hierarchies is very difficult if not impossible. Often, more than one underlying factor contribute when some activity is not well-placed in the hierarchy. Nevertheless, our conclusion is rather simple. We recommend that in developing an image recognition model (i) data collection should be prioritized for activities that are not well-placed in the hierarchy and (ii) enhancements to the label set should be made considering the potential similarity to those activities such as including more activities from the same domain. Likewise, it is important to ensure a rich variety and quantity also for the dataset that is used in calculating the similarities between the activities for hierarchical clustering. In this way, the performance of the proposed hierarchical labeling system can be improved and become more practical for real-world tasks. Besides, the predictive accuracy might improve not only at upper levels of hierarchy but also at the level of individual activities which is desirable in any prediction use case in computer vision.

\section{Conclusion and Future Work}

In this paper, we have proposed a method to automatically generate hierarchical activity trees to understand and improve how a HAR machine thinks. We have demonstrated that such a hierarchical labeling system can be constructed based on the similarity patterns in the prediction results from a state-of-the-art DL-based HAR model.
We have performed a series of experiments to prove the feasibility of labeling systems that can automatically generate hierarchical taxonomies of human activities from the machine’s perspective.

The proposed labeling technique can preserve privacy at upper levels while maintaining a better accuracy at those desired and sufficient levels for the specific use cases. 
Also, deep learning video processing is generally built on one or a few datasets, with fixed label sets. Due to the nature of supervised learning, the model can only predict behaviors that exist in its training set. Even with an unseen input, a model will produce the most likely prediction from the known behaviors, which could be wrong. Through our approach, this unseen input will be at least associated with the correct higher level labels.

The proposed approach provides tools to evaluate the model’s intelligence while keeping the human in the loop and reveals the bias of the underlying model.
Such information can also provide guidelines for collecting training data, e.g., by focusing on the activities that are not intuitively similar but clustered together by the machine-generated tree. 
The contradictions in the summarized level labels between the understandings of the machine and the human, we can refer them as cognition outliers, can reveal the underlying barriers that are preventing a machine to think intelligently like a human. 
Future research may use not only the output from the last neural network layer but also the last few layers to further investigate the machine's presentations for the similarities and distinctions between activities. If we can understand machine thinking on a deeper level, the better we can train it to truly think like a human. Such methods would also be contributing significantly to the research stream of Explainable AI. 
We also motivate researchers to further develop model architectures that can directly produce predicting outputs in a hierarchical tree structure. 
Lastly, such machine-generated hierarchies can be used in other research problems where similar hierarchies are created manually.

\bibliographystyle{unsrt}  
\bibliography{references}

\EOD

\end{document}